# Dr. Watson type Artificial Intellect (AI) Systems

Authors: Saveli Goldberg(1), Stanislav Belyaev(2), Vladimir Sluchak

((1) MGH Radiation Oncology Department, (2) Estern New Mexico Medical Center)

**Abstract**

The article proposes a new type of AI system that does not give solutions directly but rather points toward it, friendly prompting the user with questions and adjusting messages. Models of AI – human collaboration can be deduced from the classic literary example of interaction between Mr. Holmes and Dr. Watson from the stories by Conan Doyle, where the highly qualified expert Mr. Holmes, answers questions posed by Dr. Watson. Here Mr. Holmes, with his rule-based calculations, logic and memory management apparently plays the role of an AI system and Dr. Watson is the user. Looking into the same Holmes-Watson interaction, we find and promote another model in which the AI behaves like Dr. Watson, who, by asking questions and acting in a particular way, helps Holmes (the AI user) to make the right decisions. We call the systems based on this principle "Dr.Watson-type systems". The article describes the properties of such systems and introduces two particular - Patient Management System for intensive care physicians and Data Error Prevention System.

# 1 Introduction

The increasing use of Artificial Intelligence (AI) in various fields and particularly in medicine moves the problems of machine-human interaction to a higher level. Business leaders and academics are warning that current advances in AI may have major adverse consequences to society [1]:

> *"Humans, limited by slow biological evolution, couldn't compete and would be superseded by AI"*—Stephen Hawking in BBC interview, 2014.
> *AI is our "biggest existential threat,"* Elon Musk at Massachusetts Institute of Technology during an interview at the AeroAstro Centennial Symposium (2014).
> *"I am in the camp that is concerned about super intelligence*." Bill Gates (2015) wrote in an Ask Me Anything interview on the Reddit networking site.

Hazards posed by AI in the future may include:
  -Increased unemployment caused by the substitution of humans by the machines
  -Psychological problems caused by working in an environment with increased automation
  -Loss of human skills due to AI superiority
  -Possible use of AI for destructive tasks.
Those are possible future dangers, but the introduction of AI systems already runs into socio-psychological problems like distrust and rejection from specialists and loss of professional skills (learned helplessness) caused by the transfer of decision-making responsibility.

Computer ethics originated from Norbert Winner's works, - investigates the possibility of establishing ethical principles for the machines. It is hoped that human-friendly ethical AI can avoid most of these problems [2],[3]. Several approaches to the implementation of moral agents in AI are considered: formal logical and mathematical ethical reasoning [4], [5], machine learning methods based on examples of ethical and unethical behavior [6], providing AI systems with internal models to make them self-conscious [7].

Medical decisions are challenging; they involve setting goals for the outcome, collecting data, interacting with multiple entities participating in patient care, developing care plans, modifying them according to the patient's response and changing treatment goals for specific patients. Those decisions are always influenced by socio-economic factors, such as the level of technological development of the society, cultural and religious preferences, economics, various aspects of the legal and financial systems. With a fast and accelerating rate of changes, it is more and more difficult for a human to consider all these factors; the machine can be more effective here. On the other hand, a decision based solely on computer algorithms will hardly consider all ethical and cultural differences among patients. Moreover, psychological conflicts can arise within the AI -provider-patient triangle.

Modern basic artificial intelligence systems function as pure conscious minds. In essence, these systems are sociopathic. This is one of the reasons why people feel anxious and even fearful when it comes to further development of this kind of systems.

We believe that humanly reasonable explanations of AI decisions will break down the wall of mistrust and, particularly in the medical area, will make AI more considerate of the specific needs of care providers and their patients.

# 2 Principles of Dr. Watson-type systems

Can AI instead of presenting the final solution, produce some "doubts" and questions directing the user toward the solution and making him a confident author?

Few models of AI–human collaboration can be deduced from the classic literary example of interaction among Mr. Holmes, his brother Mycroft and Dr. Watson from the stories by Conan Doyle, where the highly qualified expert Mr. Holmes, answers any questions from Dr. Watson, while teaching him deductive thinking. Here Mr. Holmes, with his rule-based calculations, logic, and memory management apparently plays the role of an AI system and Dr. Watson is the user. Mycroft plays the role of an intellectual superpower, which Holmes – the main decision-maker uses rarely and reluctantly. Holmes – Mycroft relations can be another model of human-AI interaction. But here below, we propose and describe yet another one – more psychologically comfortable for a human in which the AI behaves like Dr. Watson, who, by asking questions and acting in a particular way, helps Holmes (the AI user) to make the right decisions.

Essentially Watson-Holmes's collaboration represents harmonious interaction of different aspects of human mind. Dr. Watson functioning in the background, running errands for Mr. Holmes, and asking questions represents an unconscious mind. Mr. Holmes with his brilliance, logic, and decisiveness is a representation of the conscious aspect of mind. Mycroft with his wisdom and out-of-this-world excellence represents the superconscious. Each character played their own role at the same time, when circumstances dictated, they would assume responsibilities of each other

It is known that IBM gave the name of Dr. Watson to its excellent AI system [8], but we find its style closer to Mycroft than to Dr. Watson.  Then, what are Dr. Watson characteristics that attracts Sherlock Holmes and helps him to solve problems in psychological comfort? These are precisely the ones we consider the most important for our implementation of friendly AI (and the ones that Mycroft was lacking):

  1. Dr. Watson is Holmes's chronicler. He documents Mr. Holmes' victories and failures.  He admires Holmes and rejoices at his successes.
  2. Dr. Watson asks questions every time he misses the connection in Holmes's path to the solution, thus stimulating thinking and promoting Holmes's creativity.
  3. Dr. Watson does not compete but rather collaborates with Holms. Therefore, he does not present a threat to Mr. Holmes
  4. Dr. Watson covers Holmes in dangerous situations, but even under those circumstances, he does not take over the cases.
  5. Holmes trusts Watson and does not expect him to create problems.
  6. Collaboration between Holmes and Dr. Watson resembles a game in which both play their parts enthusiastically.
  7. Holmes feels comfortable with Watson.

Let us call an AI system "Dr. Watson type system" [9],[10],[11] if it:
  -Does not offer a specific solution but rather analyses a user's solution in a way that
  would reveal logical inconsistencies or insufficient data for such a solution (Fig.1)
  -Uses game situations, psychological techniques, visual presentation of information to enhance and
  stimulate the intellectual activity of the user.

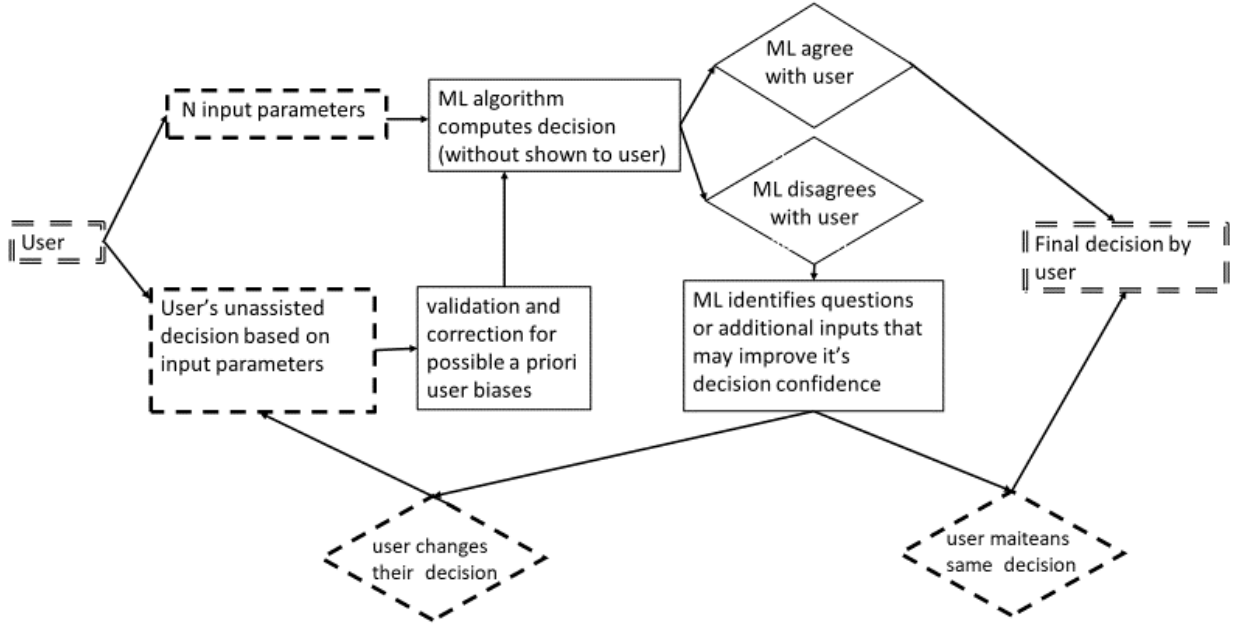

Fig1.

Let us translate listed above features of Dr Watson styles into the requirements to the AI system:

1. The system should document and analyze the dynamics of the development of the expert/specialist. For example the variability in the accuracy of their predictions from patient to patient, how the accuracy of their decisions changes over time, and how their performance compares with the average results in the specialty and/or a certain subgroup of specialists.
2. The system should discover contradictions and omissions on the path to the solution and prompt the user with corresponding questions, answering to which would either confirm the decision or raise doubts and direct the user towards another one.
3. The interaction process should start from formalization of the outcome and user's plan of care, and then – the system should influence the decision-making using organization of the data and asking clarifying questions.
4. The system should provide alarms in situations of "obvious" errors – hence it should have a dictionary of those errors.
5. The confidentiality of specialist-machine communication must be ensured. It is advisable to delete the documentation of their interactions and interim solutions.
6. The system should provide context-based reference and help. It is highly desirable to include game elements in the interaction between the expert and the system.
7. The system should have a convenient, efficient, and friendly user interface.

Evaluating and refining a user's decision, without directly providing them with an AI solution, is fundamental for Dr. Watson type systems. We base this approach on algorithms for local explanation of the AI solutions. If AI's decision matched the user's decision, then the factors that were not suitable for such a decision are presented to the user as inappropriate. If AI made a decision that differs from the user's decision, then the factors that had been most useful for the AI's decision are offered to them as factors that contradict the user's decision.

# 3. Dr. Watson's type system formalization

This comes down to the formal explanation on how the system generates questions based on the decision provided by the user and parameters that led him to that decision.

Let $\mathbf{x} = [x_1, x_2, .., x_n]$ be a vector of the n input parameters to the algorithm. $x_i$ can be continuous (numerical) or categorical (Boolean) variable. Let X be a set of $\mathbf{x}$.

Let $\mathbf{D} = \{\alpha_j\}$, j=1,...,k be the set of k possible decisions. For each solution $\alpha_1,..., \alpha_K$, one should have a typical representative $\mathbf{s}(\alpha_j)=[s_i(\alpha_j), s_{i+1}(\alpha_j), ...., s_n(\alpha_j)] \in \mathbf{X}$, which is either the center of gravity of the training sample for $\alpha_i$ or a set of medians for each coordinate in such a sample or, in the absence of a training sample, it is represented by experts. Let us know $\mathbf{v} = [v_1,...., v_{i-1}]$, i-1 <= n, and the expert's decision $\alpha_{Holmes}$. First, let us check if additional questions are needed to confirm $\alpha_{Holmes}$. To do this, we randomly select values for coordinates i,i+1,...,n. If there is a point $[v_1,...., v_{i-1},t_i,..,t_n] \in \mathbf{X}$, where the solution will be $\alpha_{Watson} \neq \alpha_{Holmes}$, then it makes sense to ask questions and remember $\alpha_{Watson}$. If there are several options for an alternative solution to $\alpha_{Watson}$ with a random selection of the values of coordinates i,i+1,..., n, then the most common solution is retained.

Consider the vector $[v_1,...., v_{i-1}, s_i(\alpha_{Holmes}), s_{i+1}(\alpha_{Holmes}), ...., s_n(\alpha_{Holmes})] \in \mathbf{X}$. Let us apply the local explanation of the $\alpha_U$ solution for the vector $[v_1,...., v_{i-1}, s_i(\alpha_{Holmes}), s_{i+1}(\alpha_{Holmes}), ...., s_n(\alpha_{Holmes})]$. The most important parameter (coordinate) from this explanation i, i + 1, ..., n may be our question. If there is a ector $[v_1,...., v_{i-1},t_i,..,t_n] \in \mathbf{X}$ with the conclusion $\alpha_{Watson}$, then we apply a local explanation of the solution $\alpha_{Watson}$ to it. Now, for each parameter, we can summarize its significance for $\alpha_{AI}$ and $\alpha_U$. Based on this the system asks a question. If AI makes for $[v_1,...., v_{i-1}, s_i(\alpha_{Holmes}), s_{i+1}(\alpha_{Holmes}), ...., s_n(\alpha_{Holmes})]$ a decision $\alpha_{Watson} \neq \alpha_{Holmes}$, then a clarifying question is asked on the most significant parameter for the local explanations of the $\alpha_{Watson}$ solution from the whole spectrum of parameters from 1 to n. Any posteriori, model-independent, and local interpretation methods are good for such tasks [12].

# 4 Patient Management System

This section describes SAGE patient management system (PMS) as an implementation of Dr. Watson type AI.

## 4.1 Requirements and subsystems.

The requirements to SAGE PMS were combined from those seven listed above as fundamental for Watson-type systems and from the results of the interviews with experts who worked with another PMS – a predecessor of SAGE – DINAR2 [13],[14]. Few intensive care providers with no less than two years of experience with DINAR2 were asked to rate possible features of a hypothetically perfect PMS system.

Features were evaluated on a 5-point scale. The top 3 were added to the requirements: assessment of treatment adequacy; assessment of treatment efficiency and assessment of the course of disease.

It was presumed that a physician would oversee the management of multiple patients, while the mid-level would be responsible for the procedures, bedside patients care and data entry. The remaining data would be imported automatically from the network. Essentially the software would be integrated with EHR. SAGE consists of 6 interconnected subsystems: "Information import", "Diagnostics", "Treatment efficiency", "Treatment Adequacy", "Integral assessment of the patients", "Discontinuation of observation" [15]

## 4.2 Information import

The goal of this subsystem is collecting information on the patients (Fig 2). The subsystem consists of two sections. In the first section a physician determines input parameters (vital signs, labs, imaging) and frequency of their measurements for a specific patient. The second section is an interface for data input (manual and automatic). SAGE would check the data entry for possible errors.

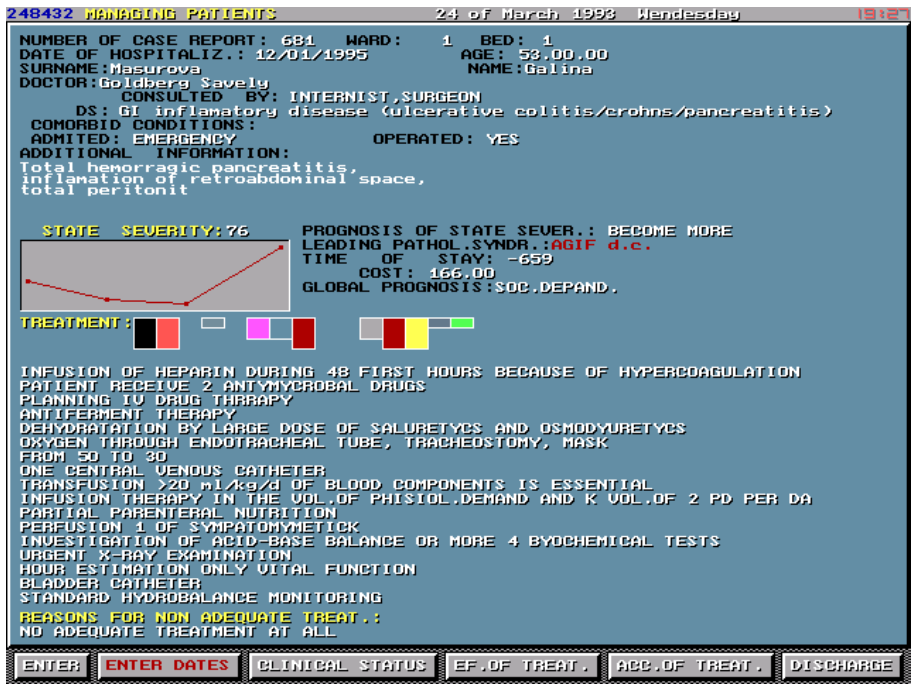

Fig 2

## 4.3 Diagnostics.

The goal of this subsystem is assistance with selection of a diagnostic hypothesis about pathologic syndromes. Matching information regarding the patient's condition with the physician's diagnostic hypothesis, SAGe offers its own view on the outcome of a particular patient.

SAGE decision-making process is based on a set of rules derived from the opinions of several experts, the rule set was created during "Diagnostic Game "[16],[17] similar to the one used for its predecessor DINAR 2. The initial set of rules, which had been verified and tested for over 5 years, have been transplanted from DINAR 2. Major technical difference between "diagnostic games" for SAGe and DINAR 2 is in the use of databases.

Here how this subsystem works. If SAGe prognosis matches physician's opinion, SAGe does not manifest itself in any way. If the assessment of the physician and SAGe do not match, SAGe displays the specific parameters which do not match. At this point the physician is prompted with an additional question. SAGe determines the most useful question that would challenge the physician's opinion. If an answer to this question matches SAGe hypothesis, but the physician's original assessment of the situation remains unaltered, the SAGe asks a new question (maximum two additional questions). Since the procedure for making a diagnosis is organized as a movement along a decision tree, the algorithm for choosing a clarifying question is relatively simple. The ease of finding an additional question was one of the reasons for choosing decision tree construction as a machine learning algorithm.

Visualization of information plays an important role in the diagnostic subsystem (Fig 3). Therefore, in order to simultaneously present various quantitative and qualitative parameters, it was decided to use unified scales for those parameters.

Now we would like to emphasize an important concept implemented in SAGe. It is normalization of different quantitative and qualitative parameters like BP, presence of infiltrates on CXR, HR, and WBC. For a provider working with the system SAGe displays those parameters in a conventional way (for most laboratory tests, the "normal range" is defined as values falling within 2 std of the mean), but internally the software operates with normalized data.

Here is how the normalization is done:

Different degrees of "abnormality" are determined based on parameter limits established by experts.

Parameter normalization $x_i^{norm}$ for each **i**, based on four thresholds: $a1_i$, $a2_i$, $a3_i$, $a4_i$ that are defined by a group of expert physicians.

$x_i < a1_i$ : strong deviation $x_i^{norm} = 0 + x_i / a1_i$

$a1_i <= x_i < a2_i$ : abnormal: $x_i^{norm} = 1 + (x_i - a1_i)/(a2_i - a1_i)$

$a2_i <= x_i < a3_i$ : normal: $x_i^{norm} = 2 + (x_i - a2_i)/(a3_i - a2_i)$

$a3_i <= x_i < a4_i$ : abnormal: $x_i^{norm} = 3 + (x_i - a3_i)/(a4_i - a3_i)$

$a4_i <= x_i$ : strong deviation: $x_i^{norm} = 3 + x_i /(a4_i)$

Thus, normalized parameters will belong to five intervals: [0,1), [1,2), [2,3), [3,4),

and the interval of numbers greater than 4.

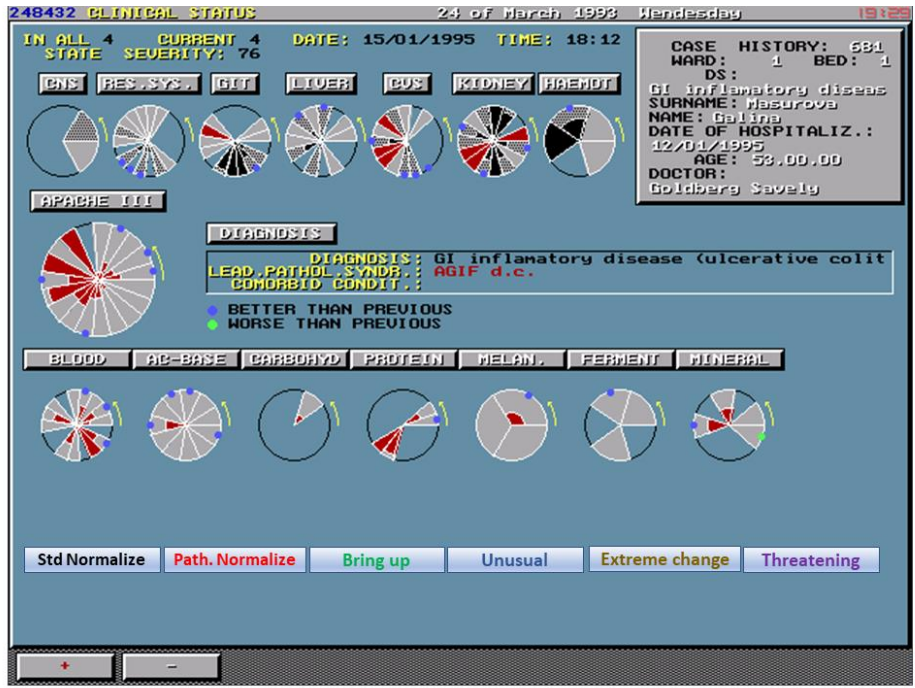

Fig 3.

To visualize unique conditions of each patient we used pie-charts with sections presenting parameters and color boundaries within section presenting their numeric values. The GUI displays available parameters, shows contradictions and omissions (Fig 4). It helps in assessing specific cases by generating links with similar cases from the past.

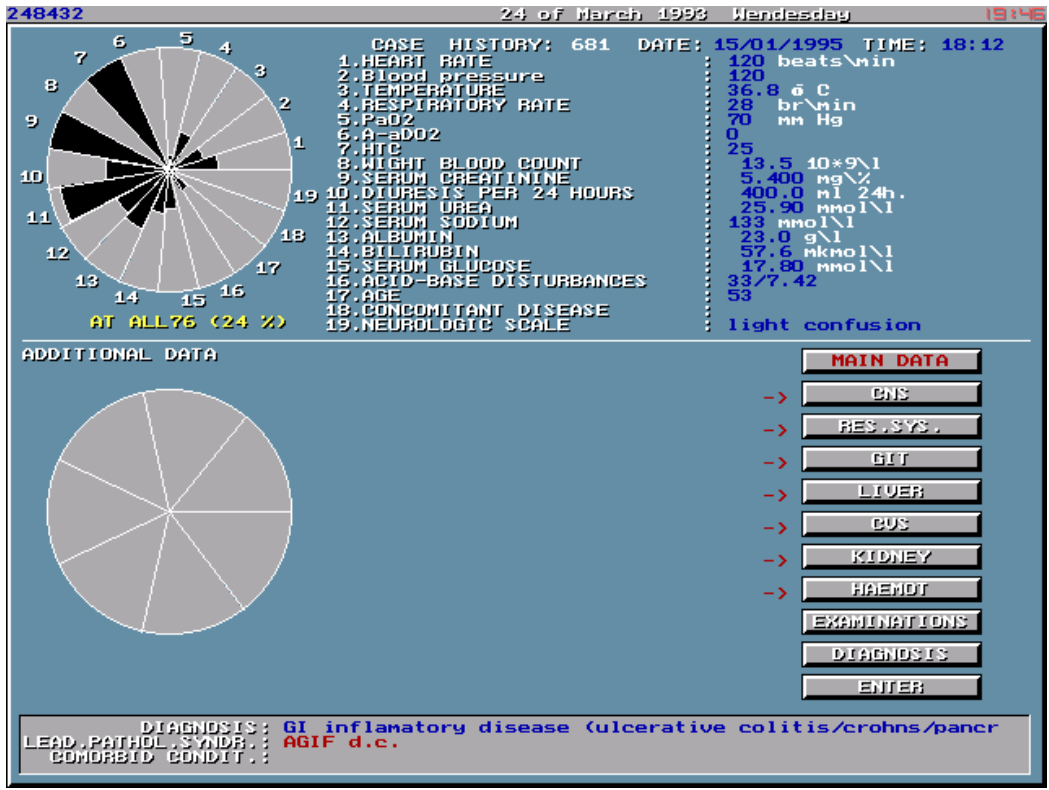

Fig 4

In this subsystem each organ is displayed as a circle divided into several sectors corresponding to the most important parameters. The intensity of color on each sector corresponds to the values of the normalized parameters. The numeric value in those sectors is a conventional representation of those parameters. Each combination of the parameters reflect the integral condition of a patient and the system GUI draws corresponding patterns on the pie-charts. With regular use of the system, the user gets trained to recognize those visual patterns and associate them with the conditions of a patient.

The provider can access historical data, including trends.

In the starting windows of SAGe GUI, the user has a choice from 6 items (buttons).

1st -brings up changes in any of the parameters over a given period.

2nd - allows direct access to the most important parameters of the vital organs.

3rd -gives access to the parameters exhibiting unusual patterns. For example, typically HR increases with rise of body temperature. The system would classify falling temperature and increasing HR combination as unusual. The typical correlations among different parameters have been built in the system.

4th -displays parameters with the most extreme dynamics (for example, sudden and significant change in HR).

5th - presents parameters with the most threatening dynamics (for example, sustained critically elevated BP). The 6[th] - allows a provider to view the parameters that they have viewed under similar circumstances in the past.

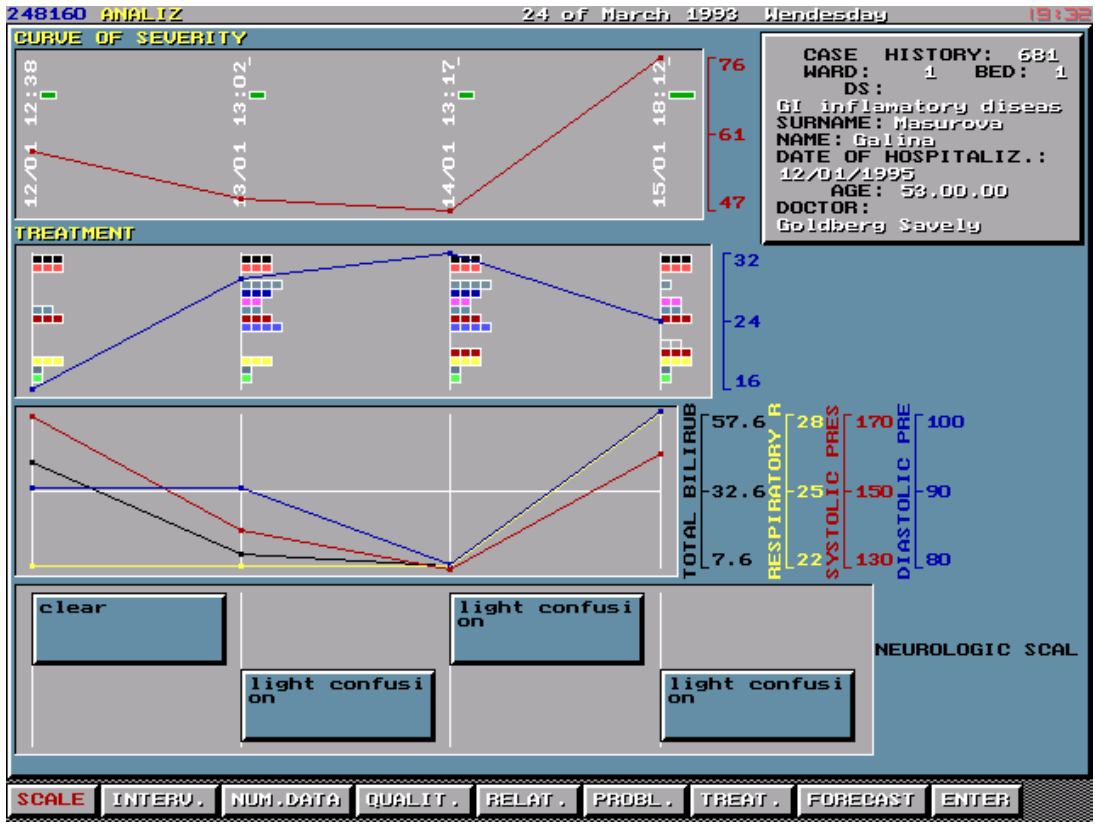

Fig 5

This screen can show no more than 4 quantitative and 2 qualitative parameters at a time. Ther0efore, only the most significant parameters from the groups represented by buttons 2, 3, and 4 are displayed. The clinical significance of these parameters is estimated by SAGe based on normalized data (Fig 5). The same style of dimitic parameters presentation is used by Shklovsky-Kordi N in his clinical applications [18]

When a new parameter is entered, one of the previously displayed parameters (the least important one) disappears. All parameters are shown in conventional units, although there is a way to review those parameters in their normalized form, which could be useful in some instances.

Once the leading syndrome/diagnosis is determined, the normal and abnormal values for different parameters are recalculated by the system the way that they would most closely match that specific syndrome/diagnosis.

Matching the information on the patient's condition and the physician's diagnostic hypothesis, SAGe generates its own diagnostic hypothesis. If it matches the physician's opinion, SAGe does not manifest itself in any way otherwise, SAGe displays the mismatching parameters. At this point the physician is prompted with an additional question. SAGe determines the most useful question that would challenge the physician's opinion. If an answer to this question matches SAGe hypothesis, but the physician's original assessment of the situation remains unaltered, the SAGe asks a new question (maximum two additional questions).

For example, a patient with sepsis is being treated with fluids and antibiotics. Persistently elevated HR, despite adequate resuscitation, would be brought to the provider's attention, as it might be a manifestation of another undiagnosed condition. In this case the system may ask the question: "Are you expecting the HR to remain this high?"

## 4.4 Treatment effectiveness.

To assess the effectiveness of treatment, we used a multifactorial evaluation of patient response to the therapy. This SAGe subsystem displays the trends of the overall severity of the patient's condition, the severity of leading syndromes relative to the dynamics of type and intensity of treatment.

We understood that:
- Overall changes in the patient's condition may be caused by the factors that have not been analyzed by the system.
- Despite an overall improvement of the patient's condition, an unfavorable trend of some parameters can lead to serious problems in the future.

In this regard, a provider can compare dynamics of overall severity of patient with changes of any quantitative and qualitative parameters.

At the same time, SAGe offers trends of parameters that need attention when defining treatment. These parameters are selected by the SAGe algorithm.

SAGe presents five groups of such parameters:

  1 group - parameters reflecting a malfunction of organ systems presently not affected (presumably) by the disease

  2 group - parameters with a paradoxical dynamic (those that significantly deviate from overall trend)

  3 group - parameters exhibiting the most significant fluctuation from the baseline

  4 group - the most life-threatening parameters

  5 group - parameters reviewed by the user under similar conditions in the past

For each parameter from the groups, 1 through 5 SAGe assigns a rank from 1 to 7 depending upon its importance. The user is presented with 4 quantitative and 2 qualitative parameters, which are ranked the highest. When too many parameters have the same rank, preferences are formed according to the group numbers.

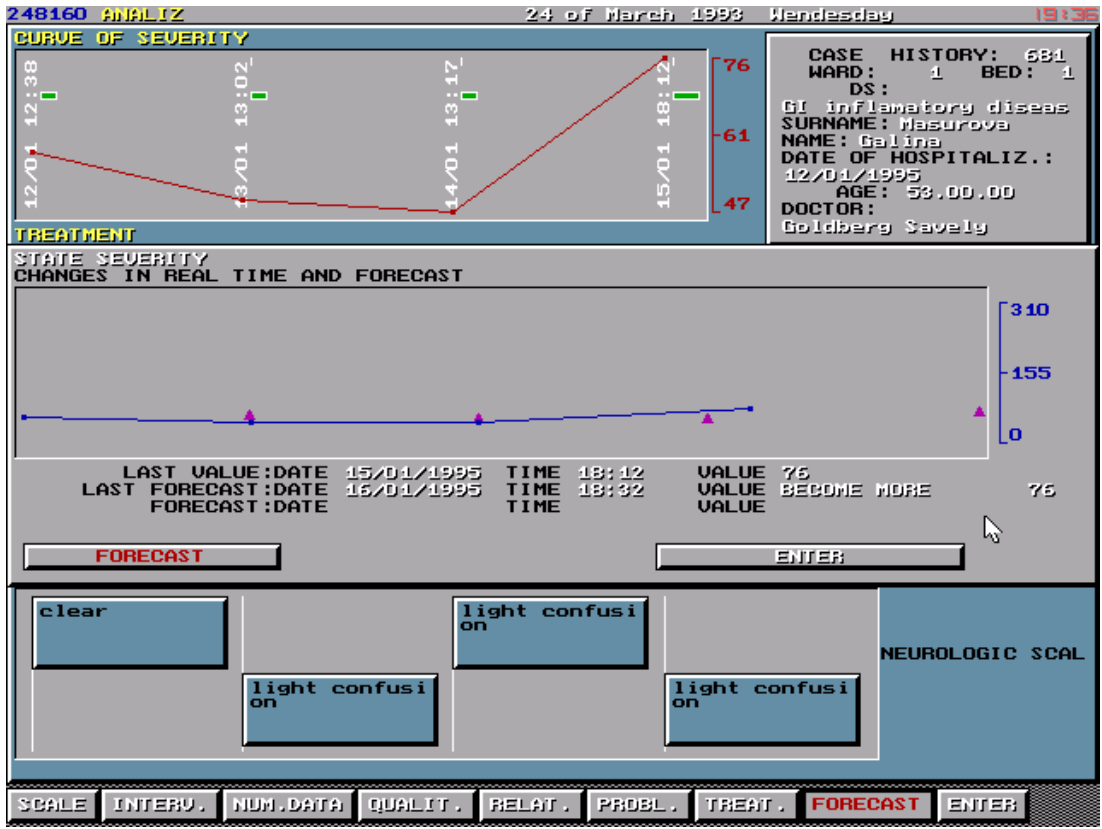

Fig 6

A provider is required to make an independent prognosis regarding the course of future events in response to treatment. In case of discrepancy between prognosis and reality, a provider is required to explain probable causes of treatment ineffectiveness and deterioration of the patient's condition (Fig 6).

Our prior experience of using similar questions in DINAR 2 revealed their great stimulating effect on the optimization of the plan of care.

## 4.5 Treatment adequacy.

We view "treatment adequacy" as a balance between intensity of treatment and the desirable response of the patient to that treatment.

Therapy can be excessive. For example, massive fluid resuscitation and amputation of an extremity would be extreme and excessive interventions for superficial infection of a leg. Therapy can be ineffective due to limited understanding of the nature of the disease or information pertinent to a specific patient. Regardless of the circumstances the provider must be conscious of the consequences of the chosen plan of care.

This subsystem incorporates multiple factors like admission data, initial plan of care and its subsequent modifications, response of the patient to treatment, provider's prognosis, and type of therapy.

All data is presented in a form of the growth of the flower. Each element of the flower represents a particular aspect of a process of taking care of the patient. The root is HPI, the stem is the trend of the severity of the patient's condition, the branches are unrealized prognoses, the flower is the current values for physiological parameters (similar to the circle from "Diagnostics") , the leaves are leading syndromes, props and strings are the types of treatment, the atmospheric phenomena are reasons for changes of the patient's condition, the  types of soil and problems with props and strings are reasons for ineffectiveness of treatment and its adequacy and so on (Fig 7).

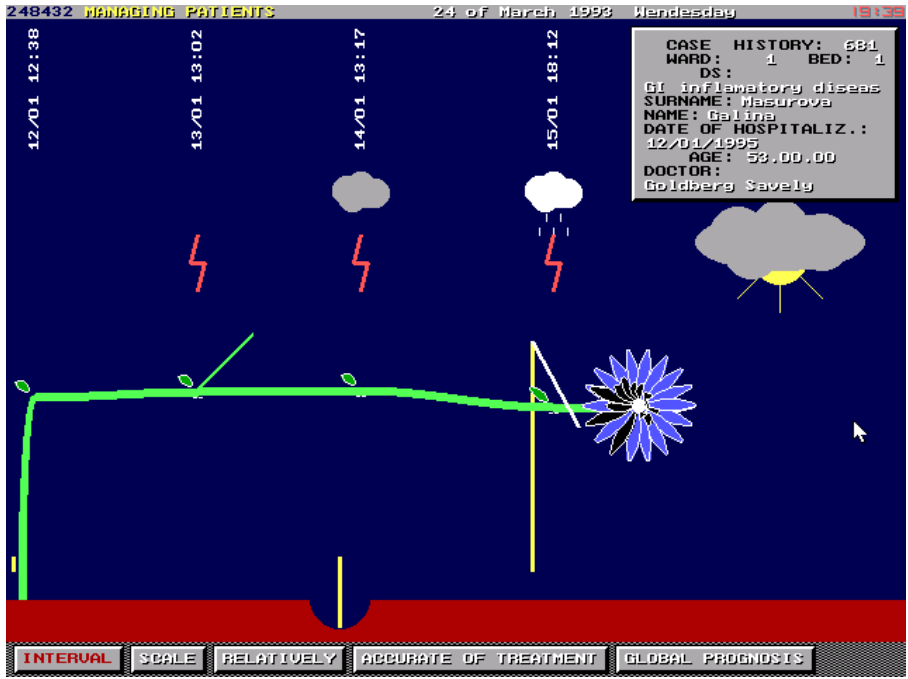

Fig 7

The system detects and analyzes contradictions in the provider's decision-making process, as it compares the provider's prognosis with the outcomes which are reflected in improvement or deterioration of the patient's condition.

Analyzing the provider's input and the patient's condition at various points SAGe tries to assess the degree of controllability of the therapeutic process and subsequently displays it as a particular color of the picture's background.

By distorting the image of a flower, SAGE brings to attention various contradictions in the management of the patient.

## 4.6 Taking a patient off the observation

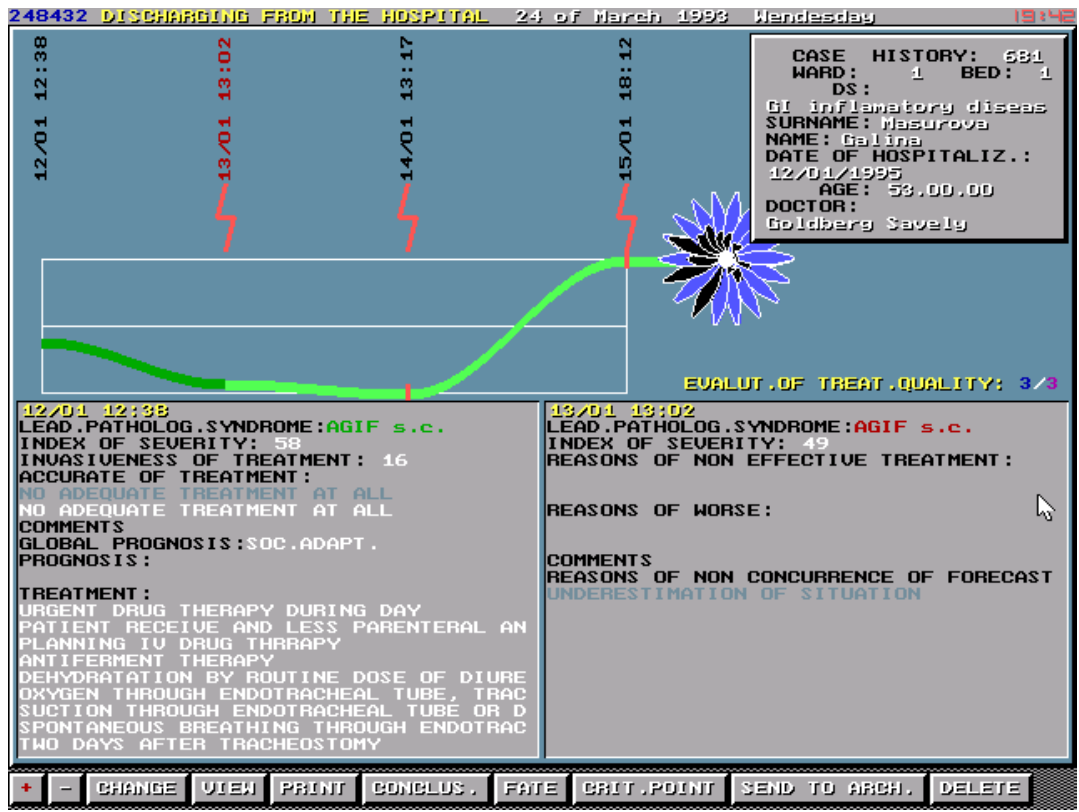

Fig 8

In this subsystem, the physician finalizes the case. The goal of this system is a retrospective analysis of the case by a provider. It is done when the actual unfolding of the events is known, the problems of the patient's treatment have been explained in a new, more accurate way, and contradictions in the provider's earlier reflections have been eliminated (Fig 8).  Missing interventions are determined as well as the ones that from today's positions were harmful or useless. Critical points in the progression of the disease are identified and explained. These explanations and these breakpoints remain in the patient file in the SAGE database. The explanations for the mismatch of the real-time predictions and their outcome based on the retrospective analysis done at the time when the patient was "removed from observation" are stored in the database without reference to a particular patient. We felt that this feature would incentivize a more thorough approach to post hoc analysis and reduce the potential danger of altering history records.

## 4.7 Integral assessment of patients in the department.

This subsystem starts and completes the work with SAGe. The landing screen of this subsystem presents the department plan with wards and medical beds. Each bed is displayed as a window with patient information retrieved from other subsystems. The leaders among patients are defined by the severity of state, unfavorable state dynamics, financial expenses on treatment, medical complexity of the case, etcetera (Fig 9).

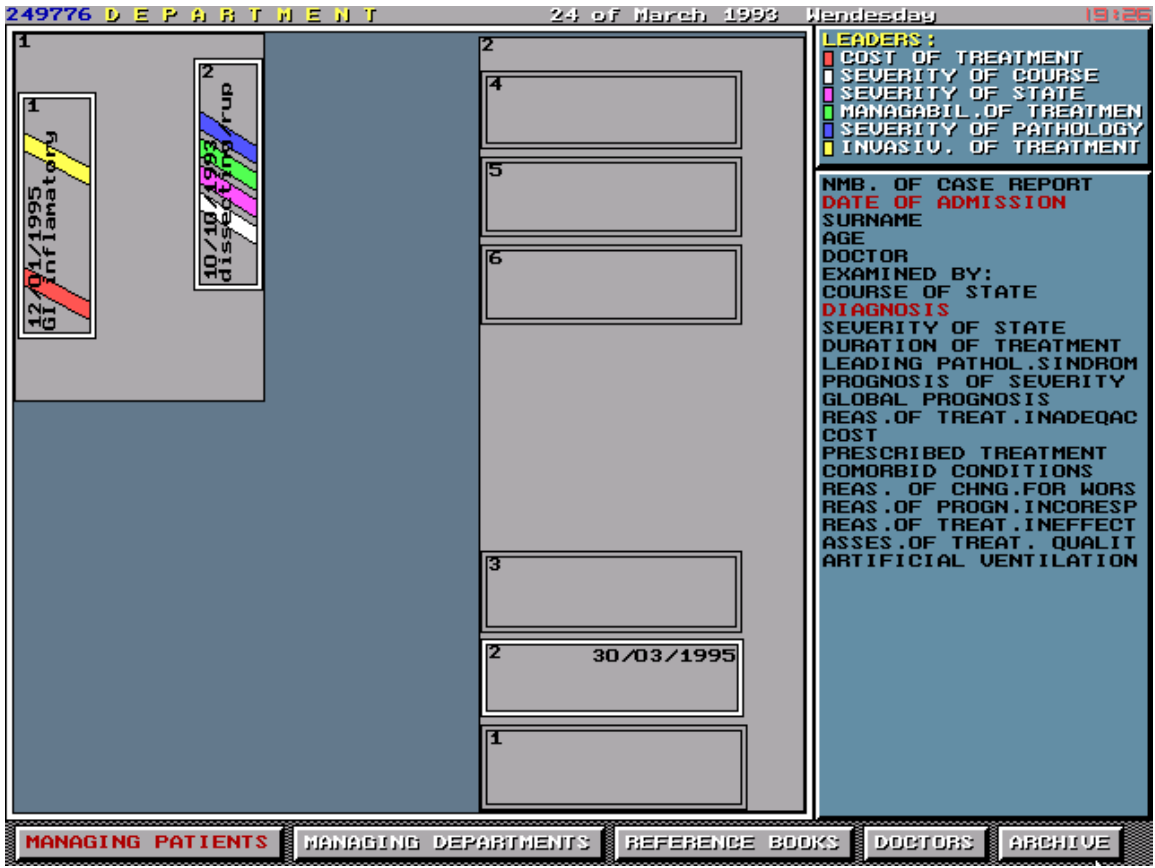

Fig 9.

The leading positions are derived from comparison of the following indexes:

N1 (t,i) - index of controllability of the treatment process for patient i at moment t

N1 (i,t) = F1( t-1,i ) + F2( t,i ) + F3( t,i )

Where F1( t-1,i ) – discrepancy between prognosis and reality at moment t

F2( t,i ) – loss of control of a patient i at moment t

F3( t,i ) - lack of coordination in the treatment revealed at moment t

N2(t,i) – index of the unfavorable condition of the state dynamics at moment t

N2(t,i)= ( I( t,i ) - I( t - 1 ) ) * ln I( t -1)

Where I (t,i) - the physiological assessment of the severity [ 10 ] at moment t

N3(t,i) - index of treatment invasiveness [11] at moment t

A physician has full access to integral characteristics and reports on his patients and partial access to the data on patients treated by other physicians.

## 4.8 Features of Dr. Watson type system presented in SAGe.

Basic principles of Dr.Watson-type systems were formulated while conceiving SAGE. However, we found that DINAR2 already had some of them. At the development stage of SAGE, we introduced subsystems and their principles [19].

Here below are some details of the implementation of Dr. Watson-type AI in different subsystems.

### 4.8.1 Discovery of contradictions and omissions.

In the "Diagnostics" subsystem, it is provided by revealing information that does not fit into the diagnostic hypothesis of a physician.
In the "Efficiency"- by revealing disparity between prognosis and reality.
In the "Adequacy" - with showing contradictions in the physician's reflections.
In "Taking a patient off observation" - by prompting the physician to create an archive of his own posteriori speculations and conclusions about the treatment course of the patients.

### 4.8.2 Attempts to direct a physician toward alternative solutions.

- In the "Diagnostics"- introduced with the help of directing questions.
- In the "Efficiency" - with showing signs demanding attention.
- In the "Adequacy" - with showing possible mistakes in previous treatment.
- In all the subsystems, algorithms make their choice of signs necessary to stimulate the physician, based on the final tasks of the subsystem accepted by SAGe. Algorithms developed for SAGe possess an apparatus of self-tuning to the user.

### 4.8.3 Encouragement and motivation.

- In "Integral assessment" – provided by making the case history of a patient, possibility to receive integral information and scope of all patients at once, archiving data and convenient access to the archive.
- In the "Diagnostics" – with access to standard assessment of the severity of a patient's state, control of the authenticity of the information being entered by an assistant into case history.
- In the "Efficiency" – with a comparison of trends in various parameters and evaluation of the therapy invasiveness.
- In the "Adequacy" – with creating an archive of the physician's reflections and problems arising during treatment of a patient.

Information about a person's activity entered SAGe is accessible only to this person. The input of SAGe is protected with passwords. Data entry into SAGe (done by the physician's assistant) takes place in a special input subsystem.

# 5. Data Error Prevention system.

There are database errors that cannot be detected using expert rules or numerical methods,
for example – one of the most dangerous - error in class definition occurring while forming a training sample for AI. As we know, each error of this type in machine training generates a set of operational errors.

The purpose of the proposed system is to identify and prevent such errors in the data entry process.

At the heart of this system, we placed a pre-shipment summary function. Such a summary is standard for many modern electronic data entry systems; for example, online stores or banks offer the user to review the entered information before completing the transaction. The AI system "Follow-up Summary" (originally called the Summary Page [19]) generates case-specific screens allowing the user to notice inconsistencies in the data and to suspect possible errors in the record.
A complete view of the data is not possible for clinical trial databases, typically containing tens, if not hundreds, of fields spanning multiple screens. The Follow-up summary is designed to mitigate this problem by applying the following principles:

- focus on key areas
- the visual combination of the fields to facilitate the detection of inconsistencies
- indication of probable errors

## 5.1 The "Follow-up Summary" page consists of the following main sections:

**1) Verbatim list of selected record fields that represent key demographic and clinical information.**

The presentation of these fields is organized to show inconsistencies in the data. This is achieved through a menu of items of data, organizing the order in which data is presented, highlighting, and changing the font for key parameters. The combinations of parameters that are suspicious of internal incompatibility with each other were obtained in advance based on expert rules and the existing database.

*Consider an example*:
The analytical algorithm suggests that information about cancer recurrence may have been missed because the patient's record has a large tumor size, a positive margin as a result of surgery, and the second resection was performed relatively late after lumpectomy. The average relapse rate in the first 5 years after diagnosis is about 4%. If the tumor is> 2 cm, the positive margin and the distance between the lumpectomy and the second resection is more than 1 month, then the probability of recurrence within a year between the lumpectomy and the second resection increases to 10%. However, an alarm cannot be triggered at 90% false positives.

Regular presentation data without problems should be:

*Female 55 years old, T-stage 2, N-stage 0, M stage 0. Tumor size 2.2 cm, CR negative, PR negative, tumor size 2.2 cm. Biopsy 02/28/2009, Lumpectomy 03/04/2009, positive margin. Local relapse 12.06.2009. 2nd Resection 01.07.2009, negative margin. No RT, No Chemotherapy*

Case presentation with suspected non-relapse:

*Female 55 years old, **Tumor size 2.2 cm,** T stage 2, N stage 0, M stage 0. Tumor size 2.2 cm, PR negative. 02/28/2009 Biopsy. No RT, No Chemotherapy, 03/04/2009 **lumpectomy with a positive margin. After 8 months, the 2nd resection.** <u>No relapses.</u>*

2) **Schematic representation of the chronology of the patient's clinical course**.

A single view without problems:

_02/28/2009___03/04/2009__________________12/06/2009____01/07/2009

  Biopsy         Lumpectomy                           Local Recurrence  2nd Resection

Case presentation with suspected non-relapse or 2nd Resection date:

_02/28/2009___03/04/2009______________________________<span style="color:red">01/07/2009</span>

Biopsy         Lumpectomy                                            2nd Resection

3) **"Possible Error" section that lists likely errors based on a set of rules that take into account all fields in the record. Typically, these errors, while likely (> 50%), are not fully certain, and therefore do not merit an interruptive alert**.

An example of a possible error that can be identified in the breast cancer database is a new surgical procedure (e.g., modified radical mastectomy after the original lumpectomy) without a documented relapse of cancer (Fig. 10). This is most likely an error where the information about the relapse was omitted. However, the error is not definite since it is also possible that the patient herself requested a more aggressive procedure to be reassured that cancer will not recur. Most of the rules for identifying possible errors involve data I" tems entered on different screens of the database, so it can be difficult for users to combine them. The "*Follow-up Summary*" performs this function.

The additional benefit of using "*Follow-up summary*" is to view the dangerous spot in the data. As shown above, errors often occur next to each other. A user who finds and corrects one error or omission in a recording may also be more likely to view the remainder of the recording and find other errors.

The "*Follow-up summary*" was implemented as a single screen in a Microsoft Access database that could be accessed from anywhere in the recording. Users could access the "*Follow-up summary*" at any time during the data entry process or ignore the "*Follow-up summary*".

"Follow-up summary" was implemented as a single screen in a Microsoft Access database. Users can access the Follow-up summary page at any time during the data entry process if they so choose.

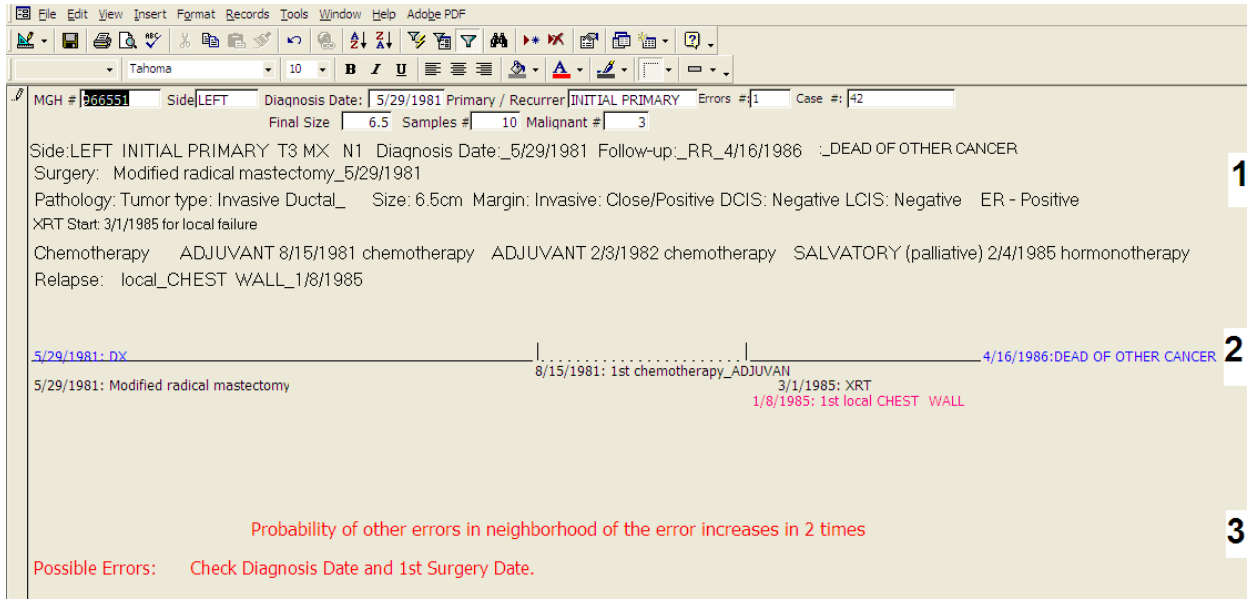

Fig 10.

## 5.2 The "Follow-up summary" implementation.

A working implementation of "*Follow-up Summary*" uses the rules that are based mainly on expert opinion. But potentially suspicious parameter combinations can also be generated using machine learning algorithms. For this case, we used a promising method, "*Minimal Antisyndrome*" [20].

Here is the definition of the "minimal antisyndrome":

"minimal antisyndrome" of the class $A \subset X$ is the minimum set of symptoms that does not occur in the elements of class **A**. For example, of all emergency calls, 35% were for men and 10% for pregnant patients. Combination "*pregnant patient*: Yes" plus "*gender*: Male" will be the minimal antisyndrome of the class of ambulance patients. The minimal antisyndrome of class **A** plus any other parameter is an antisyndrome of class **A**, but if we exclude any parameter from the minimal antisyndrome, then the combination of other remaining parameters is not an antisyndrome for class **A**.

With a set of minimal antisyndromes, it is possible to construct a recognition rule for class **A**.

If the probability of occurrence of a set $x_1, .., x_m$ (m<<n), provided that the parameters are independent, is $p(x_1, .., x_k) = p(x_1) * ... * p(x_k)$ and if in reality in class the frequency of the set $(x_1, .., x_k)$ was much less than $(p(x_1) * ... * p(x_k))$ then set of $x_1+x_2+..+x_k$ is candidate to be minimal antisyndrome.

## 5.3 "Follow-up summary" utilization

To determine the utilization of the "*Follow-up summary*", we analyzed all 1356 records (200 new and 1156 updated) that were entered or updated in database B between 07/04/2008 and 03/04/2009. Distribution of "*Follow-up summary*" utilization per record is presented in Table 2, Fig 11. "*Follow-up summary*" was accessed in slightly less than half (44.2%) of all records. Most commonly, it was accessed only once per record but in 2.4% of the records, it was accessed three or more times. "*Follow-up summary*" was accessed for 69.0% of entries of new records but only for 40.0% of the updates of the records already in the database (p=0.01). Three data entry technicians worked with the database during this period. Their "*Follow-up summary*" access rate was 84.6%, 47.1% and 11.8%, respectively (p<0.001).

In multivariable analysis the odds of "*Follow-up summary*" access was 12% higher if the patient's cancer had T stage greater than 1 and over 50% higher if this was a new record rather than an existing one being updated. Patients' vital status, tumor grade, whether the cancer had relapsed, and the year of diagnosis were not significantly associated with "*Follow-up summary*" access (Fig 12).

**Table 2 - "*Follow-up summary*" Utilization per Record**

| *"Follow-up summary"* accesses per record | Records, N (%) |
|---|---|
| 0 | 757(55.83%) |
| 1 | 492(36.28%) |
| 2 | 75(5.53%) |
| 3 | 15(1.11%) |
| 4 | 10(0.74%) |
| 5 | 2(0.15%) |
| 6 | 2(0.15%) |
| 7 | 2(0.15%) |
| 8 | 1(07%) |

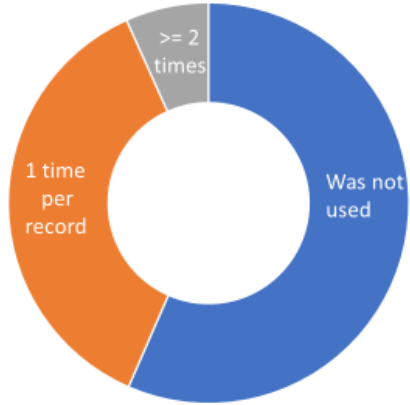

Fig 11

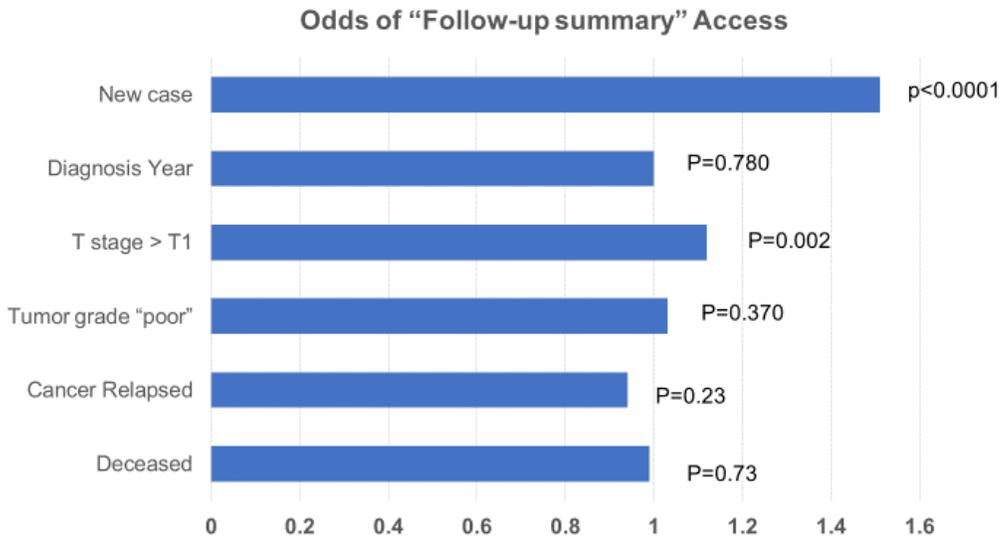

Fig 12.

## 5.4 "Follow-up summary" effectiveness

To evaluate the effectiveness of the "*Follow-up summary"* we analyzed the same 1356 records as we used to test utilization of the "*Follow-up summary"*. Among these, 164 had documented a remission but had the final vital status "Died from Disease", indicating that a relapse had occurred. Five of these 164 records (3.05%) did not have any information in the *Date of Relapse* field recorded. "*Follow-up summary"* was accessed for 86 of the 164 records. None of the records where "*Follow-up summary"* was accessed had a missing date of relapse (p = 0.023).

    Among the 1,156 records that were updated during the study period, "*Follow-up summary"* was accessed for 462 records. Relapse information was entered after "*Follow-up summary*" access for 16 (3.5%) of the records where "*Follow-up summary*" was utilized and for 4 (0.6%) of the remaining records (p = 0.0003). In all 16 records where relapse information was entered after "*Follow-up summary*" access, it was done within 10 minutes of accessing the "*Follow-up summary"*. Similarly, out of the 9 records where the date of the very first relapse was entered, 7 records (p = 0.034) were done within 10 minutes after "*Follow-up summary"* access. In multivariable analysis, the odds of entry of date of relapse increased by 3.8% if the patient's vital status was "Deceased" and by 4.5% if the "*Follow-up summary"* was accessed during the record update. Tumor grade, tumor stage, and the year of cancer diagnosis were not significantly associated with the probability of entry of date of relapse. There was no significant difference in the probability of entry of a relapse date among individual technicians (Fig 13).

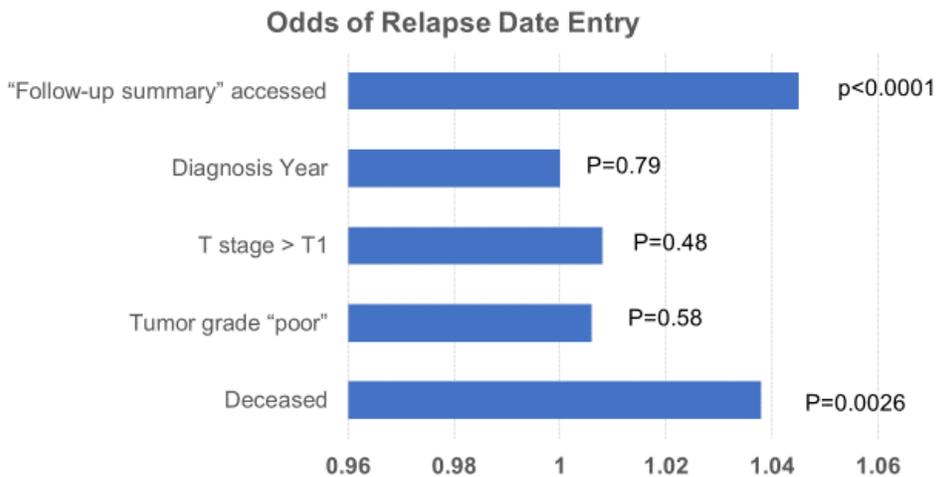

Fig 13

# 6 Conclusion.

We have proposed Dr. Watson type AI system and identified it as the one that:
  -Does not directly offer a specific solution but instead analyses a user's solution and
  reveals logical inconsistencies and missing data.
  -Uses game situations, psychological techniques, visual presentation of information to enhance and
  stimulate the intellectual activity of the user.
We defined technical features of this type of AI systems as follows
  1. The system should document and analyze the dynamics of the development of the user. For example, the variability in the accuracy of their predictions from patient to patient, how the accuracy of their decisions changes over time, and how their performance compares with the average results in the specialty and/or a certain subgroup of specialists.
  2. The system should discover contradictions and omissions on the path to the solution and prompt the user with corresponding questions, answers to which would either confirm the decision or raise doubts and direct the user towards another one.
  3. The interaction process should start from the formalization of the user's plans and possible outcomes. Then – the system should influence the decision-making using the organization of the data and asking clarifying questions.
  4. The system should provide alarms in situations of critical errors – hence it should have a dictionary of those errors.
  5. The confidentiality of specialist-machine communication must be ensured.
  6. The system should provide context-based reference and help. It should include game elements in the interaction with the user.
  7. The system should have a convenient, efficient, and friendly user interface.

Elements of Dr. WTS type system have been used in DINAR2 - EPR Patient Management software of the Regional Pediatric Advisory Center. We think that the implementation of those elements contributed to the success of DINAR2.

We introduced Data Error Prevention System with Follow-up summaries as another implementation of the Dr. Watson type AI system. This system is currently in use at MGH.

We developed and tested SAGE – the most complete "Dr. Watson type system" as an AI tool for managing patients in one of the intensive care units of Russia.

# Reference

1. Jim Torresen. A Review of Future and Ethical Perspectives of Robotics and AI, Front. Robot. AI, 15 January 2018 | https://doi.org/10.3389/frobt.2017.00075
2. Anderson, M., and Anderson, S. L. (2011). *Machine Ethics*. New York: Cambridge University Press
3. Wallach, W., and Allen, C. (2009). *Moral Machines: Teaching Robots Right from Wrong*. New York: Oxford University Press.

4. Dennis, L. A., Fisher, M., and Winfield, A. F. T. (2015). Towards verifiably ethical robot behaviour. *CoRR* abs/1504.03592. Availble at: http://arxiv.org/abs/1504.03592
5. Arkin, R. C., Ulam, P., and Duncan, B. (2009). *An Ethical Governor for Constraining Lethal Action in an Autonomous System*. Technical Report GIT-GVU-09-02.
6. Deng, B. (2015). Machine ethics: the robot's dilemma. *Nature* 523, 20–22. doi:10.1038/523024a
7. Winfield, A. F. (2014). "Robots with internal models: a route to self-aware and hence safer robots," in *The Computer after Me: Awareness and Self-Awareness in Autonomic Systems*, 1st Edn, ed. J. Pitt (London: Imperial College Press), 237–252.
8. Mohamed Nooman Ahmed; Andeep S. Toor; Kelsey O'Neil Cognitive Computing and the Future of Health Care Cognitive Computing and the Future of Healthcare: The Cognitive Power of IBM Watson Has the Potential to Transform Global Personalized Medicine. IEEE Pulse, Volume: 8 Issue: 3, 2017 https://ieeexplore.ieee.org/abstract/document/7929430
9. Goldberg S.I., Meshalkin L.D. A New Class of AI-Systems ( DrWT-Systems ). Technical Cybernetics. 5 , 1992, p. 217-223
10. Goldberg S, Meshalkin L. Assisting Dr.WT - Systems. In  East - West Conference on Artificial Intelligence: From Theory  to Practice, Moscow, Russia, 1993, p.207-209.
11. SI Goldberg,  Inference Engine the Systems of the Dr. Watson Type - DIMACS Workshop, Rutgers University, 1997
12. Christoph Molnar. Interpretable Machine Learning. A Guide for Making Black Box Models Explainable, 2019-12-17 https://christophm.github.io/interpretable-ml-book/
13. SI. Goldberg, V.L. Lomovskikch, E.K. Tsibulkin. Computer aided intellectual environment for Re-gional Pediatric Consultatine Centre on Intensive Therapy / // Information Technology Serving General Practice: Exanting Health Care Horizons. - Melboum, Australia.- 1993.- P. 232-241.
14. Kazakov DP. EMERGENCY PEDIATRIC CARE IN LARGE REGION, Ekaterinburs,UPSS, 2004 (in Russian)
15. Goldberg S. I., Tsibulkin E. K., Rudnov V. A. et al. // Computer program SAGE as a tool for standardized assessment of resuscitation patients. // Anesthesiology i Reanimatology, 1997, no. 1, p. 11-15. (in Russian)
16. I.Gelfand, B.Rosenfeld, M.Shifrin, Essays on collaboration of mathematicians and physicians (2nd ed.), Moscow, URSS, 2004 (in Russian).
17. M.A.Shifrin, O.B.Belousova, E.I. Kasparova. Diagnostic games, a Tool for Clinical Experience Formalization in Interactive "Physician - IT-specialist" Framework". Proceedings of the Twentieth IEEE International Symposium on Computer-Based Medical Systems, 2007, 15-20.
18. Shklovsky-Kordi N, Zingerman B, Rivkind N, Goldberg S, Davis S,  Varticovski L, Krol M, Kremenetzkaia A, Vorobiev A, Serebriyskiy I, Computerized Case History, History - an Effective Tool for Management of Patients and Clinical Trials. In Connecting Medical Informatics and Bio-Informatics R. Engelbrecht et al. (Eds.) ENMI, 2005 p 53-57.
19. Saveli I Goldberg, Andrzej Niemierko, Maria Shubina , Alexander Turchin "*Summary Page*": a novel tool that reduces omitted data in research databases, *BMC Medical Research Methodology* 91 (2010)
20. Goldberg S. Diagnostics on the basis of the informative space of the antisyndromes. *Problems of Control and Information Theory.* 1984;13(6): pp 401-411.